# Feature Level Fusion of Face and Fingerprint Biometrics

A. Rattani, D. R. Kisku, M. Bicego, *Member, IEEE* and M. Tistarelli

*Abstract*— The aim of this paper is to study the fusion at feature extraction level for face and fingerprint biometrics. The proposed approach is based on the fusion of the two traits by extracting independent feature pointsets from the two modalities, and making the two pointsets compatible for concatenation. Moreover, to handle the 'problem of curse of dimensionality', the feature pointsets are properly reduced in dimension. Different feature reduction techniques are implemented, prior and after the feature pointsets fusion, and the results are duly recorded. The fused feature pointset for the database and the query face and fingerprint images are matched using techniques based on either the point pattern matching, or the Delaunay triangulation. Comparative experiments are conducted on chimeric and real databases, to assess the actual advantage of the fusion performed at the feature extraction level, in comparison to the matching score level.

## I. INTRODUCTION

In recent years, biometric authentication has seen considerable improvement in reliability and accuracy, with some of the traits offering good performance. However none of the biometrics are 100% accurate. Multibiometric systems [1] remove some of the drawbacks of the uni-biometric systems by grouping the multiple sources of information. These systems utilize more than one physiological or behavioral characteristic for enrollment and verification/ identification. Ross and Jain [2] have presented an overview of Multimodal Biometrics with various levels of fusion, namely, sensor level, feature level, matching score level and decision level.

However it has been observed that, a biometric system that integrates information at an earlier stage of processing is expected to provide more accurate results than the systems that integrate information at a later stage, because of the availability of more richer information. Since the feature set contains much richer information on the source data than the matching score or the output decision of a matcher, fusion at the feature level is expected to provide better recognition performances.

Fusion at matching score, rank and decision levels have been extensively studied in the literature [3][4]. Despite the abundance of research papers related to multimodal biometrics, fusion at feature level is a relatively understudied problem. As a general comment, it is noticed that fusion at feature level is relatively difficult to achieve in practice because multiple modalities may have incompatible feature sets and the correspondence among different feature spaces may be unknown. Moreover, concatenated feature set may lead to the problem of curse of dimensionality: a very complex matcher may be required and the concatenated feature vector may contain noisy or redundant data, thus leading to a decrease in the performance of the classifier [5]. Therefore, in this context, the state of the art is relatively poor.

Ross and Govindarajan [5] proposed a method for the fusion of hand and face biometrics at feature extraction level. Gyaourova et al. [6] fused IR-based face recognition with visible based face recognition at feature level, reporting a substantial improvement in recognition performance as compared to matching individual sensor modalities. Recently, Ziou and Bhanu [7] proposed a multibiometric system based on the fusion of face features with gait features at feature level.

Even though face and fingerprint represent the most widely used and accepted biometric traits[1], no methods for feature level fusion of these modalities have been proposed in the literature. The possible reason is the radically different nature of face and fingerprint images: a face is processed as a pictorial image (holistic approach) or as composed by patches (local analysis), while fingerprint is typically represented by minutiae points. In this paper a recently introduced methodology for face modeling [8] is exploited, which is based on a point-wise representation of a face called *Scale Invariant Features Transform* (SIFT), thus making the feature

Manuscript received May 25, 2007. This work has been partially supported by grants from the Italian Ministry of Research, the Ministry of Foreign Affairs and the Biosecure European Network of Excellence.

A. Rattani is with DIEE, University of Cagliari, Piazza d'Armi, 09123 Cagliari – Italy( Phone: +39 348 5579308   Fax:  +39 070 6755782 email: ajita.rattani@diee.unica.it).

D. R. Kisku is with DCSE, Indian Institute of Technology Kanpur, 208016 Kanpur – India (email: drkisku@iitk.ac.in).

M. Bicego is with DEIR, University of Sassari, via Torre Tonda 34, 07100 Sassari – Italy (e-mail: bicego@uniss.it).

M. Tistarelli is with DAP, University of Sassari, piazza Duomo 6, 07041 Alghero (SS) – Italy (Phone: +39 079 9720410, Fax: +39 079 9720420 e-mail: tista@uniss.it).

---

[1] Remarkably face and fingerprint data has been adopted as biometric traits to be included in the European electronic passport. The same traits are currently being used at the US immigration for manual identification of passengers.

level fusion of face and fingerprints possible.

Thus, this paper proposes a novel approach to fuse face and fingerprint biometrics at feature extraction level. The improvement obtained applying the feature level fusion is presented over score level fusion technique. Experimental results on real and chimeric databases are reported, confirming the validity of the proposed approach in comparison to fusion at score level.

## II. FACE AND FINGERPRINT BIOMETRICS

### A. Face Recognition based on Scale Invariant Feature Transform Features (SIFT)

The face recognition system, preliminary introduced in [8], is based on the SIFT [9] features extracted from images of the query and database face. The SIFT features represent a compact representation of the local gray level structure, invariant to image scaling, translation, and rotation, and partially invariant to illumination changes and affine or 3D projections. SIFT has emerged as a very powerful image descriptor and its employment for face analysis and recognition was systematically investigated in [8] where the matching was performed using three techniques: (a) minimum pair distance, (b) matching eyes and mouth, and (c) matching on a regular grid. The present system considers spatial, orientation and keypoint descriptor information of each extracted SIFT point. Thus the input to the present system is the face image and the output is the set of extracted SIFT features $s=(s_1, s_2,.......s_m)$ where each feature point $s_i=(x, y, \theta, k)$ consist of the $(x, y)$ spatial location, the local orientation $\theta$ and $k$ is the keydescriptor of size $1x128$.

### B. Fingerprint Verification based on Minutiae matching technique

The fingerprint recognition module has been developed using minutiae based technique where fingerprint image is normalized, preprocessed using Gabor filters, binarized and thinned, is then subjected to minutiae extraction as detailed in [10]. However to achieve rotation invariance the following procedure is followed in the image segmentation module.

In order to obtain rotation invariance, the fingerprint image is processed by first detecting the left, top and right edges of the foreground. The overall slope of the foreground is computed by fitting a straight line to each edge by linear regression. The left and right edges, which are expected to be roughly vertical, are fitted with lines of the form $x = my + b$ and for the top edge the form $y = mx + b$ is applied. The overall slope is determined as the average of the slopes of the left-edge line, the right-edge line, and a line perpendicular to the top edge line. A rectangle is fitted to the segmented region and rotated with the same angle to nullify the effect of rotation. Although the method is based on the detection of edges, only a rough estimate of the fingerprint boundaries is required for fitting the lines and extracting the edges. This improves the robustness to noise in the acquired fingerprint image. The input to this system is a fingerprint image and the output is the set of extracted minutiae $m=(m_1, m_2,.......m_m)$, where each feature point $m_i=(x, y, \theta)$ consist of the spatial location $(x, y)$ and the local orientation $\theta$.

## III. FEATURE LEVEL FUSION SCHEME

The feature level fusion is realized by simply concatenating the feature points obtained from different sources of information. The concatenated feature pointset has better discrimination power than the individual feature vectors. The concatenation procedure is described in the following sections.

### A. Feature set compatibility and normalization

In order to be concatenated, the feature pointsets must be compatible. The minutiae feature pointset is made compatible with the SIFT feature pointset by making it rotation and translation invariant and introducing the keypoint descriptor, carrying the local information, around the minutiae position.

The local region around each minutiae point is convolved with a bank of Gabor filters with eight different equally spaced degrees of orientation *(0, 22.5, 45, 67.5, 90, 112.5, 135, and 157.5)*, eight different scales and two phases *(0 and π/2)*, giving a keydescriptor of size *1x128*. The rotation invariance is handled during the preprocessing step and the translation invariance is handled by registering the database image with the query images using a reference point location [11]. Scale invariance is achieved by using the dpi specification of the sensors. The keypoint descriptors of each face and fingerprint points are then normalized using the *min-max* normalization technique ($s_{norm}$ and $m_{norm}$), to scale all the *128* values of each keypoint descriptor within the range *0 to 1*. This normalization also allows to apply the same threshold on the face and fingerprint keypoint descriptors, when the corresponding pair of points are found for matching the fused pointsets of database and query face and fingerprint images.

### B. Feature Reduction and Concatenation

The feature level fusion is performed by concatenating the two feature pointsets. This results in a fused feature pointset $concat=(s_{1norm}, s_{2norm},...s_{mnorm},....m_{1norm}, m_{2norm}, m_{mnorm})$. Feature reduction strategy to eliminate irrelevant features can be applied either before [7] or after [5-6] feature concatenation.

### C. Feature Reduction techniques

*1. K-means clustering.* The normalized feature pointsets ($s_{norm}$ and $m_{norm}$) are first concatenated together (concat). Redundant features are then removed using the "*k-means*" clustering techniques [12] on the fused pointset of an individual retaining only the centroid of the points from each cluster. These clusters are formed using spatial and orientation information of a point. The keypoint descriptor of each cluster's centroid is the average of keypoint descriptors of all the points in each cluster. The distance classifier used is euclidean distance. The number of clusters are determined using the PBM cluster validity index [13]. Since, the feature poinset from the two modalities i.e., face and fingerprint are affine invariant and moreover, they are normalized using normalization technique as discussed before. They are treated simply as a set of points belonging to an individual irrespective of whether they are extracted from face or fingerprint thus making *K-means* clustering possible.

*2. Neighbourhood Elimination.* This technique is applied on the normalized pointset of face and fingerprint ($s_{norm}$ and $m_{norm}$) individually. That is, for each point of face and fingerprint, those point that lie within the neighbourhood of a certain radius (20 and 15 pixels for face and fingerprint respectively on experimental basis) are removed giving $s_{norm}$' and $m_{norm}$', the reduced face and fingerprint pointsets. Spatial information is used to determine the neighbours of each considered point. The result of neighbourhood elimination is shown in Fig. 1.

*3. Points belonging to specific regions.* Only the points belonging to specific regions of the face i.e., specific landmarks like the eyes, the nose and the mouth lower portion and the fingerprint images (the central region) are retained as reduced pointset. Face images in BANCA Database are pre-registered with respect to eyes and mouth location and the nose tip is manually identified for the current experiments. The corepoint in fingerprint is located using a reference point location algorithm discussed in [11]. A radius equal to *85 and 120* pixels was set for the face and fingerprint feature points selection as shown in Fig. 2. The SIFT points around specific landmarks on face carry highly discriminative information as experimented and reported in [8]. The region around core point accounts for combating the effect of skin elasticity and non-linear distortion due to varing pressure applied during image acquisition as it is the least affected region.

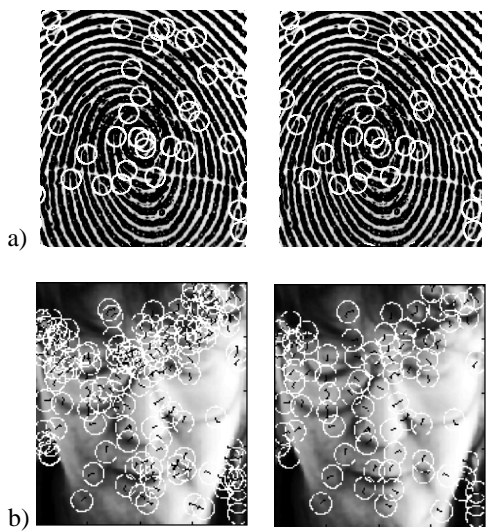

**Fig. 1.** Effects of the neighbourhood elimination on a) Fingerprint and b) Face

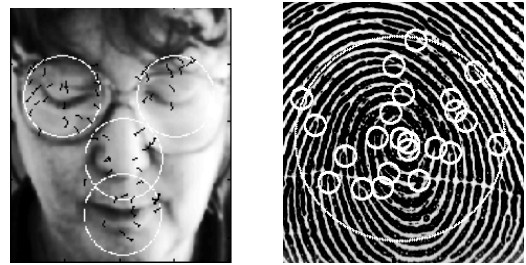

**Fig. 2** Example of selected regions on a face (left) and a fingerprint (right)

The aims of the "*k-means*" and "*neighbourhood elimination*" techniques are to remove redundant information and at the same time retaining most of the information by removing onlyl the points which are very near, as computed using euclidean distance, to a specific point. As these points may not provide any additional information because of being in vicinity. And the aim of "points belonging to specific region" is to consider only the points belonging to highly distinctive region. Thus keeping only optimal sets.

*D. Matching techniques*

The concatenated features pointset of the database and the query images *concat and concat' respectively (in which the feature reduction techniques have already been applied even before or after concatenation)* are processed by the matcher which gives matching score based on the no. of matching pairs found between the two pointsets. In this study two different matching techniques are applied.

*1. Point pattern matching.* This technique aims at finding the percentage of points "paired" between the concatenated feature pointset of the database and the query images. Two points are considered paired only if the spatial distance (1), the direction distance (2) and the Euclidean distance (3) between the corresponding key descriptors are all within some are within a pre-determined threshold, set with 4 pixels, 3°, 6 pixels for $r_o, \Theta_o, k_o$ on the basis of experiments:

$$sd(concat'_j, concat_i) = \sqrt{(x'_j - x_i)^2 + (y'_j - y_i)^2} \leq r_0 \quad (1)$$

$$dd(concat'_j, concat_i) = \min(|\theta'_j - \theta_i|, 360^o - |\theta'_j - \theta_i|) \leq \theta_o \quad (2)$$

$$kd(concat'_j, concat_i) = \sqrt{\sum_i (k^i_j - k'^i_j)^2} \leq k_0 \quad (3)$$

where the points *i* and *j* are represented by *(x, y, θ, k) with **k** = $k^1$... $k^{128}$* of the concatenated database and query pointsets *concat`* and *concat, sd* is the spatial distance, *dd* is the direction distance, and *kd* is the keypoint descriptor distance. The one to one correspondence is achieved by selecting among the candidates points lying within the threshold of spatial, direction and Euclidean distance, the one having mimimum Euclidean distance for the keypoint descriptor. Since, the feature pointsets are rotation, scale and translation invariant, in case of fingerprint, the registartion is done at image preprocessing level as explained earlier and SIFT features for face are already affine invariant features. This obviates the need to calculate transformation parameters for aligning the database and query fused pointsets.

The final matching score is calculated on the basis of the ratio of the number of matched pairs to the total number of feature points found in the database and query sets, for both monomodal traits and for the fused feature pointset.

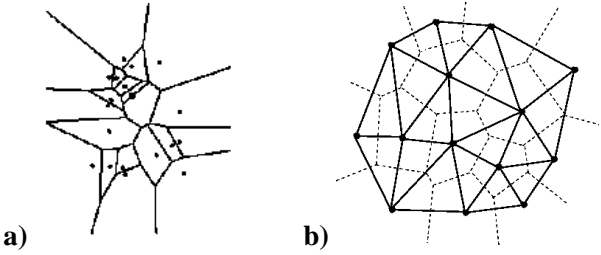

**Fig. 3.** Triangulation of pointset: a)Voronoi diagram b)Delaunay triangulation

*2. Matching using the Delaunay Triangulation technique.* In this case, instead of considering individual points, triplet of points are grouped together as new features. Given a set $S$ of points $p$, $p_2$, ..., $p_N$, the Delaunay triangulation of $S$ is obtained by first computing its *Voronoi* diagram [14] which decomposes the 2D space into regions around each point such that all the points in the region around $p_i$ are closer to $p_i$ than delaunay triangulation is computed by connecting the centers of every pair of neighboring Voronoi regions.

The Delaunay triangulation technique [15] is applied individually on the face and the fingerprint normalized pointset $s_{norm}$ and $m_{norm}$ and then on the concatenated feature pointsets $concat=(s_{norm}, m_{norm})$. Five features are computed from the extracted triplet of points. (a) The minimum and median angles ($\alpha_{min}$ $\alpha_{med}$) of each triangle (b) The triangle side ($L$) with the maximum length (c) The local orientation ($\Theta$) of the points at the triangle vertexes (d) The ratio ($l1/l2$) of the smallest side to the second smallest side of each triangle (e) The ratio ($l2/l3$) of the second smallest side to the largest side of each triangle.

All these parameters compose the feature vector $fv=(t_1, t_2 ,...,t_\#)$, where $t_i = (\alpha_{min}, \alpha_{med}, L, \Theta, l1/l2, l2/l3)$ is the triangle computed by the Delaunay triangulation. The process is repeated for the database and the query pointsets to get $fv$ and $fv'$. The final score is computed on the basis of the number of corresponding triangles found between the two feature vectors $fv$ and $fv'$. Two triangles are correctly matched if the difference between the attributes of the triangles $t_i$ and $t_i'$ are within a fixed threshold. As the fused poinset contain affine invariant and pre-normalized points thus making the application of delaunay triangulation possible.

## IV. EXPERIMENTAL RESULTS

The proposed approach has been tested on two different databases: the first consists of 50 chimeric individuals composed of 5 face and fingerprint images for each individual. The face images are taken from the controlled sessions of the BANCA Database [16] and the fingerprint images were collected by the authors. The fingerprint images were acquired using an optical sensor at 500 dpi.

The following procedure has been established for testing the mono-modal and multimodal algorithms:

**Training:** one image per person is used for enrollment in the face and fingerprint verification system; for each individual, one face-fingerprint pair is used for training the fusion classifier.

**Testing:** four remaining samples per person are used for testing, generating client scores. Impostor scores are generated by testing the client against the first sample of all other subjects. For the multimodal testing, each client is tested against the first face and fingerprint samples of the rest of the chimeric users. In total *50x4=200* client scores and *50x49=2450* imposters scores for each of the uni-modal and the multimodal systems are generated.

**Evaluation:** The best combination of feature reduction and matching strategy has been further tested on a real multimodal database acquired by the authors. The database consists of 100 individual with four face and fingerprint images per person. The first face and fingerprint combination is used for training and the rest three image pairs are used for testing, providing *100x3=300* client scores. Each individual is subject to imposter attack by ten random face and fingerprint pairs for a total of *100x10=1000* impostor scores. The experiments were conducted in four sessions recording False Acceptance Rate (*FAR*), False Rejection Rate (*FRR*) and Accuracy (which is computed at the certain threshold, *FAR* and *FRR* where the performance of the system is maximum ie., max *(1-(FAR + FRR)/2)*.

*A.* The face and the figerprint recognition systems were tested alone, without any modification in the feature sets, i.e. SIFT features *(x, y, $\Theta$ ,k)* and minutiae features *(x, y, $\Theta$)*. The matching score is computed using point pattern matching independently for face and fingerprint. The individual system performance was recorded and the results were computed for each modality as shown in table 1.

*B.* In the second session, the effect of introducing the keydescriptor around each minutiae point is examined. Once the feature sets are made compatible, the keypoint descriptors of SIFT and the minutiae points are normalized using the min-max normalization technique. The normalized feature pointsets are then concatenated and the *k-means* feature reduction strategy is applied on each fused pointset.

From the presented results (table 2), it is evident that the introduction of the keydescriptor for the fingerprints increased the recognition accuracy by 1.64%, and the feature level fusion outperformed both single modalities, as well as the score level fusion, with an increase in the accuracy of 2.64% in comparison to score level. The score level fusion is performed scores independently for face and fingerprint are computed independently for face and fingerprints which are then normalized and added using sum of scores technique.

*C.* In the third session, to remove redundant features, two feature reduction strategies are applied prior to concatenation. The matching is performed with the point pattern matching technique. From the experimental results, presented in table 3, it is evident that the application of the neighborhood removal technique does not increase the accuracy of the system. On the other hand, the reduction of points belonging to specific regions increased the recognition accuracy by 0.31%, while the FRR is dropped to 0%. Some statistics regarding the number of points retained in the fused poinsets, for all the three feature reduction techniques applied to one subject, are listed in table 4 and the performances are depicted in table 3.

**TABLE 1.** THE FAR, FRR AND ACCURACY VALUES OBTAINED FROM THE MONOMODAL TRAITS

| Algorithm | FRR(%) | FAR(%) | Accuracy |
|---|---|---|---|
| Face SIFT | 11.47 | 10.52 | 88.90 |
| Fingerprint | 7.43 | 12.19 | 90.18 |

**TABLE 2.** FAR, FRR AND ACCURACY VALUES OBTAINED FROM THE MULTIMODAL FUSION

| Algorithm | FRR (%) | FAR (%) | Accuracy |
|---|---|---|---|
| Fingerprint | 5.384 | 10.97 | 91.82 |
| (Face+Finger) score level | 5.66 | 4.78 | 94.77 |
| (Face+Finger) Feature Level | 1.98 | 3.18 | 97.41 |

**TABLE 3.** FAR, FRR AND ACCURACY VALUES FOR THE FEATURE REDUCTION TECHNIQUES

| Algorithm | FRR(%) | FAR(%) | Accuracy |
|---|---|---|---|
| Neighbourhood removal technique | 5.46 | 4.61 | 94.95 |
| Points belonging to specific regions | 0 | 4.54 | 97.72 |

**TABLE 4.** STATISTICS REGARDING THE NUMBER OF POINTS RETAINED IN THE THREE FEATURE REDUCTION TECHNIQUES I.E., K-MEANS, NEIGHBOURHOOD ELIMINATION AND POINTS BELONGING TO SPECIFIC LOCATIONS

| Algorithm | Face (SIFT) | Finger (Minutiae) | Fused pointset |
|---|---|---|---|
| The no. of Extracted features | 145 | 50 | 195 |
| K-means clustering technique | 145 | 50 | 89 |
| Neighbourhood removal technique | 73 | 25 | 98 |
| Points belonging to specific regions | 47 | 20 | 67 |

*D.* In the fourth session, the matcher based on the Delaunay triangulation of the poinsets is introduced. The reported results are computed for monomodal modalities, and multimodal fusion at matching score and feature extraction level. In the first case, all the feature points were included for triangle computation, in a second case only the reduced set of points was used. The results presented in Fig. 4, Fig. 5 and table 5, show that the application of the Delaunay triangulation enhances the performance of the face and fingerprint modalities alone by 5.05% and 0.82%, respectively. Moreover, the multimodal feature level fusion using the Delaunay triangulation outperforms all the feature level fusion experiments, with the increase in recognition accuracy of 0.35%. Finally, the combination of restricting the points to those belonging to specific regions and the Delaunay triangulation further enhanced the recognition accuracy by 0.44%.

This last configuration was further tested on the multimodal database acquired by the authors with multimodal fusion at score level and feature level. The results, presented in table 6, also demonstrate that the feature level fusion outperforms the score level fusion of 0.67%, also for the real multimodal database. The ROC curve obtained from the best strategy applied to the chimeric and the real multimodal databases is shown in Fig. 6.

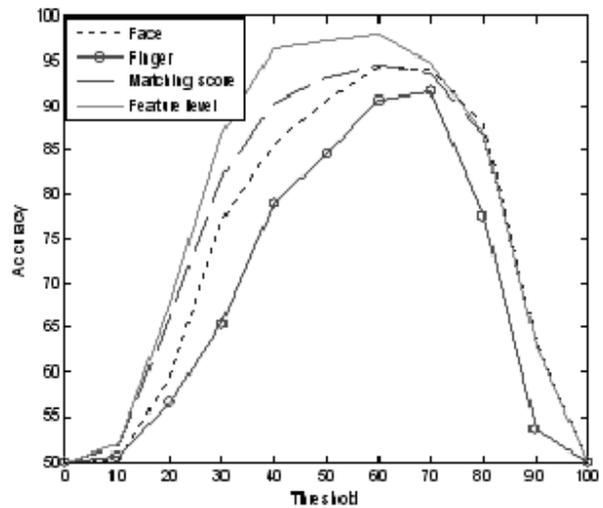

**Fig. 4.** The Accuracy Curve for Delaunay Triangulation of face, fingerprint, fusion at matching score and feature level

**TABLE 5.** FAR, FRR AND ACCURACY VALUES FOR THE DELAUNAY TRIANGULATION TECHNIQUE

| Algorithm | FRR (%) | FAR (%) | Accuracy |
|---|---|---|---|
| Face SIFT | 2.24 | 9.85 | 93.95 |
| Fingerprint | 13.63 | 3.07 | 92.64 |
| Face+Finger at Matching level | 2.95 | 8.07 | 94.48 |
| Face+Finger at Feature Level | 2.95 | 0.89 | 98.07 |
| Face+Finger at Feature level using points belonging to specific region strategy | 1.95 | 1.02 | 98.51 |

**TABLE 6.** FAR, FRR AND ACCURACY OF THE BEST MATCHING AND FEATURE REDUCTION STRATEGIES

| Algorithm | FRR(%) | FAR(%) | Accuracy |
|---|---|---|---|
| Best strategy at score fusion | 2.54 | 5.48 | 95.99 |
| Best strategy at feature fusion | 1.12 | 4.95 | 96.66 |

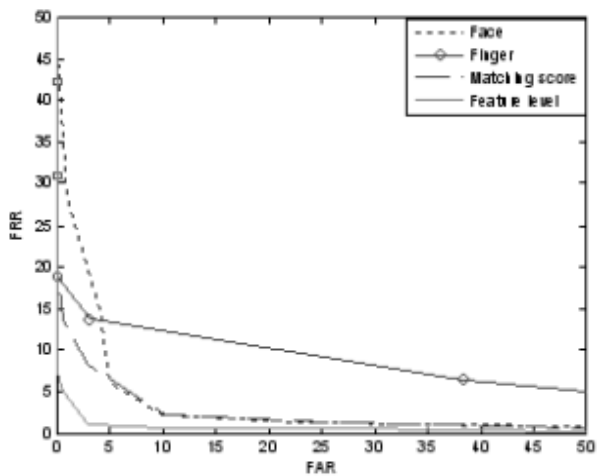

**Fig. 5.** The ROC Curve for Delaunay Triangulation of face, fingerprint, fusion at matching score and feature level

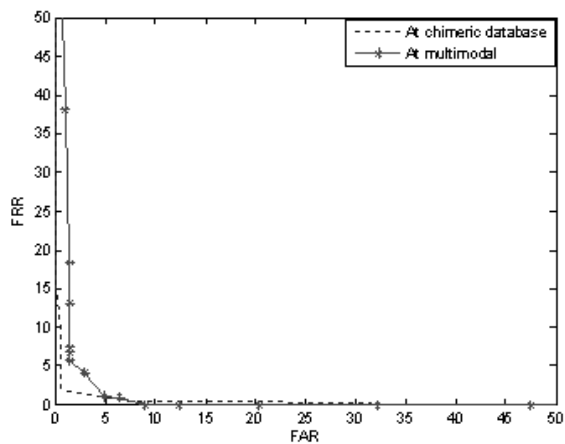

**Fig. 6.** ROC Curve for the best strategy in chimeric and multimodal database

## V. CONCLUSION

A multimodal biometric system based on the integration of face and a fingerprint traits at feature extraction level was presented. These two traits are the most widely accepted biometrics in most applications. There are also other advantages in multimodal biometric systems, including the easy of use, robustness to noise, and the availability of low-cost, off-the-shelf hardware for data acquisition.

From a system point of view, redundancy can always be exploited to improve accuracy and robustness which is achieved in many living systems as well. Human beings, for example, use several perception cues for the recognition of other living creatures. They include visual, acoustic and tactile perception. Starting from these considerations, this paper outlined the possibility to augment the verification accuracy by integrating multiple biometric traits. In this paper a novel approach has been presented where both fingerprint and face images are processed with compatible feature extraction algorithms to obtain comparable features from the raw data. The reported experimental results demonstrate remarkable improvement in the accuracies by properly fusing feature sets. This preliminary achievement, does not constitute an end in itself, but rather suggests to attempt a multimodal data fusion as early as possible in the processing pipeline. In fact, the real feasibility of this approach, in a real application scenario, may heavily depend on the physical nature of the acquired signal. The experimental results demonstrate that fusing information from independent/ uncorrelated sources (face and fingerprint) at the feature level fusion increases the performance as compared to score level. As even in the literature, it is claimed that ensemble of classifier operating on uncorrelated features increases the performance in comparison to correlated features. This work does investigation at feature level and the results are inspiring.

Further experiments, on "standard" multimodal databases, will allow to better validate the overall system performances.